# Small-sample brain mapping: sparse recovery on spatially correlated designs with randomization and clustering

**Gaël Varoquaux** GAEL.VAROQUAUX@INRIA.FR
**Alexandre Gramfort** ALEXANDRE.GRAMFORT@INRIA.FR
**Bertrand Thirion** BERTRAND.THIRION@INRIA.FR
Parietal, INRIA, NeuroSpin, bat 145, CEA, 91191 Gif-sur-Yvette France

## Abstract

Functional neuroimaging can measure the brain's response to an external stimulus. It is used to perform brain mapping: identifying from these observations the brain regions involved. This problem can be cast into a linear supervised learning task where the neuroimaging data are used as predictors for the stimulus. Brain mapping is then seen as a support recovery problem. On functional MRI (fMRI) data, this problem is particularly challenging as *i)* the number of samples is small due to limited acquisition time and *ii)* the variables are strongly correlated. We propose to overcome these difficulties using sparse regression models over new variables obtained by clustering of the original variables. The use of randomization techniques, *e.g.* bootstrap samples, and clustering of the variables improves the recovery properties of sparse methods. We demonstrate the benefit of our approach on an extensive simulation study as well as two fMRI datasets.

## 1. Introduction

Functional brain imaging, for instance using functional MRI, is nowadays central to human neuroscience research and the corresponding medical applications. Learning statistical links between the observed brain images and the corresponding subject's behavior can be formulated as a machine learning problem (Mitchell et al., 2004). Indeed, prediction from neuroimaging data has lead to impressive results, such as *brain reading*, *e.g.* guessing which image a subject is looking at from his brain activity (Haxby et al., 2001).

However, the main goal of a functional neuroimaging study is most often not prediction *per se*, but brain mapping, that is identifying the brain regions involved in the cognitive processing of an external stimuli. With fMRI, the data at hand is the brain activity amplitude measured on a grid on voxels. It is classically modeled as a linear effect driven by the cognitive task performed by the subjects. Detecting regions of the brain active in a specific task can be formalized as identifying the non-zero coefficients of a linear model predicting the external stimuli from the neuroimaging data.

From a statistical standpoint, this estimation is very challenging because of the dimensionality of the problem: the behavior must be linked to the brain activity measured on more than 20 000 voxels, often 50 000, with a few hundred observations. For this reason, the problem is most-often tackled as a mass-univariate approach: a model is fitted separately on each voxel. Detection of active voxels then suffers of a multiple comparison problem; the detection power decreases linearly with the number of variables tested. Yet, only a few brain regions are active for a given task and the linear model is sparse. There is great potential interest to use sparse recovery techniques (Carroll et al., 2009), that can recover active voxels suffering only a loss in detection power sub-linear in the number of voxels. However, with brain mapping as with many other experimental fields, the design matrix is imposed by the problem, and due to the strong correlations across regressors, univariate approaches are often more effective than multivariate approaches (Haury et al., 2011).

The main contribution of this paper is to propose an efficient sparse recovery procedure well-suited to the specificities of brain mapping situations: spatially-clustered weights in very high-dimensional correlated designs. Unlike previous work in fMRI (Carroll et al., 2009), we focus on recovery and not predictive power.

We also provide a detailed empirical study of sparse estimation in the case of large spatial configurations for the weight and correlated designs. The organization of the paper is the following: in section 2 we review results related to small sample sparse estimation, in section 3, we expose our contributed method, and in section 4 we report experimental results on spatially-correlated synthetic data and brain images.

## 2. Preliminaries: small-sample estimation of sparse linear models

The problem at hand is the recovery of sparse models in the presence of noise. Formally, given $n$ observed brain images composed of $p$ voxels, $\boldsymbol{X} \in \mathbb{R}^{n \times p}$ and the corresponding behavioral variable $\boldsymbol{y} \in \mathbb{R}^n$ related to the stimulus, the data and the target variable are related by a linear model $\boldsymbol{y} = \boldsymbol{X}\boldsymbol{\beta} + \boldsymbol{e}$ with $\boldsymbol{e} \in \mathbb{R}^n$ the measurement noise and $\boldsymbol{\beta} \in \mathbb{R}^p$ the coefficients of the model to recover. More specifically, we want to find the sparsity pattern of $\boldsymbol{\beta}$, $S = \text{supp}(\boldsymbol{\beta}) = \{i \in [1,\ldots p] \text{ s.t. } \boldsymbol{\beta}_i \neq 0\}$. For the application at hand, $n$ will typically be a few hundreds of images while $p$ can be thousands of voxels.

In general, the sample complexity –that is the number of samples requires to bound the estimation error– of a linear model with $n$ observations and $p$ variables is $\mathcal{O}(p)$. However, under certain conditions, for sparse signals, i.e. linear models for which $k \ll p$ coefficients are non-zeros, only $\mathcal{O}(k \log(p-k))$ observations are required to identify these coefficients (Wainwright, 2009; Candes et al., 2006), and the corresponding estimation problem can be solved using tractable algorithms such as $\ell_1$-penalized square loss regression –the Lasso (Tibshirani, 1994)– or greedy approaches (Tropp, 2004).

Two types of conditions govern the success of the recovery of a $k$-sparse vector from noisy measurements, i.e. model consistency of the estimator:

i) subsets of the columns of design matrix $\boldsymbol{X}$ larger than $k$ should be well conditioned, as for instance implied by the Restricted Isometry Property (RIP) (Candes et al., 2006). In particular, on the signal subspace, the design matrix $\boldsymbol{X}_S$ should be sufficiently well conditioned.

ii) the regressors on the signal subspace $\boldsymbol{X}_S$ should not be too correlated to regressors on the noise subspace $\boldsymbol{X}_{\overline{S}}$, as formalized by Tropp's Exact Recovery Condition (ERC) (Tropp, 2004) or for the $\ell_1$-penalized regression the irrepresentable condition (Zhao & Yu, 2006), or mutual incoherence (Wainwright, 2009).

In addition, the smallest non-zero coefficient of $\boldsymbol{\beta}$ must scale as the standard deviation of noise and the inverse of the minimum singular value of $\boldsymbol{X}_S$. The number of observations necessary and sufficient to almost surely recover the sparsity grows as $n_{\min} = 2\theta k \log(p-k)$ where $\theta$ depends on the various factors listed previously (Wainwright, 2009). In the specific case of $\boldsymbol{X}$ i.i.d Gaussian distributed with identity covariance, $\theta \approx 1$. Importantly, the Lasso estimator yields at most $n$ non-zero coefficients, and for $n$ below $n_{\min}$, sparse methods will often fail dramatically to recover a sparsity pattern close to the ground truth.

A common cause for failure of sparse recovery is multicolinearity: the presence of strong correlations in the design $\boldsymbol{X}$. In practice, if several columns of $\boldsymbol{X}$ are strongly correlated, sparse estimators will often select arbitrarily one of them and not the others, leading for instance to a high rate of false negatives in the support estimation if they all correspond to the signal subspace. In the extreme case of exactly collinear regressors, the Lasso, which is a non-strictly convex optimization, does not admit a unique solution. For this reason, the elastic net (Zou & Hastie, 2005) adds a strongly convex term to the Lasso in the form of a $\ell_2$ penalty. This term has the effect of grouping together correlated regressors. It opens the door to sparse recovery in relaxed conditions, in particular with ill-conditioned signal-subspace design matrices $\boldsymbol{X}_S$ (Jia & Yu, 2010). In addition, while the sample complexity of the elastic net scales similarly to that of the Lasso, the former can select more than $n$ coefficients. Another approach to ensuring strong convexity in the presence of correlated variables is based on mixed norms, imposing an $\ell_2$ penalty on sub-groups of features known a priori to covary (Yuan & Lin, 2006). When the group structure is unknown, Grave et al. (2011) proposed the trace-lasso, adapting the penalization to the design matrix. The limitation of this approach is that in the small-sample limit, the design matrix can only define a small number of strictly convex directions. For image data, correlation across neighboring voxels can be modeled using overlapping groups (Jacob et al., 2009).

Another challenge of high-dimensional sparsity recovery from noisy observations is the difficulty to control false positives, i.e. inclusion in the estimated support $\hat{S}$ of variables not present in the true support $S$. For the Lasso, theoretical bounds on false detections are conditional on choosing the right value for the parameter $\lambda$, controlling the amount of $\ell_1$ regularization. This optimum depends on the noise level and structure, and is generally considered challenging to set in the small-sample settings. For these reasons, Bach (2008) and Meinshausen & Bühlmann (2010) introduce resampled estimators based on the Lasso. As it is well known in mass-univariate analysis, such sampling of the poste-rior yields control of the probability of false selection with weaker constraints on the model, here $\lambda$. In addition, different variables in a strongly correlated group can be selected across the resampling estimates, and as a result these schemes can in theory recover correlated variables and more variables than observations. Another Lasso variant, the randomized Lasso (Meinshausen & Bühlmann, 2010) introduces a random perturbation of the design $\boldsymbol{X}$ by rescaling the regressors. It achieves sparse recovery under a RIP-like condition, bounds on sparse eigenvalues of the design, without the need of an irrepresentable-like condition imposing a strong separation between signal and noise regressors. Provided $\lambda$ is chosen large enough, the randomized lasso yields good control of the inclusion of noise variables even with a small number of observations.

While most of the theoretical results on sparse recovery have been established for the square loss, these techniques can carry over to other losses, for instance the logistic loss, which is strongly convex as the square loss, but is well suited to classification problems, rather than regression. Bach (2010) extends non-asymptotic least-square results to logistic regression by approximating it as a weighted least square and show that similar conditions on the design apply for sparse recovery with $\ell_1$ penalization. Small sample learning rates for $\ell_1$-penalized logistic regression were established earlier by Ng (2004) based on the non rotational invariance of the $\ell_1$ ball. This general argument gives a simple necessary condition for sub-linear sample complexity of empirical risk minimizers: the corresponding optimization problem should not be rotational invariant. In addition the rotational asymmetries of the learning problem should correspond to a good representation of the data. Indeed, adapting the representation of a signal such that its decomposition is more sparse is key to the performance of sparse methods.

## 3. Randomization and clustering for sparse recovery in brain imaging

Our approach to sparse recovery on spatially correlated images can be summarized by *i)* clustering highly-correlated variables before sparse regression *ii)* randomizing the clustering and sparse estimation. Our work is based on the randomized lasso (Meinshausen & Bühlmann, 2010), which we first briefly present.

**The randomized lasso** The randomized lasso consists in randomly perturbing the design matrix by taking only a fraction of the training samples and randomly scaling each variable, in our case each voxel. By repeating the later procedure and then counting how often each variable is selected across the repetitions, each variable can be assigned a score. Higher scores denote variables likely to belong to the true support.

Let $i \in [1\ldots l]$ denote the repetition and $\boldsymbol{\beta}^i$ be the corresponding estimated coefficients. The design matrix $\boldsymbol{X}^i$ is formed by a random fraction $\pi$ of the training data. Each column of $\boldsymbol{X}^i$ is then randomly scaled using a Bernoulli distribution to 1 or to $1 - \alpha$ with equal probability. The procedure is a subsampling of the data and a random perturbation of each column. The stability score of each voxel $v$ is then the percentage of the repetitions for which the voxel has a non-zero weight as estimated by sparse regression, *i.e.* is used for the prediction: $\boldsymbol{a}_v = \#\{i \text{ s.t. } v \in \text{supp}(\boldsymbol{\beta}^i)\}/l \in [0, 1]$. The estimated support is then defined as $\{v \text{ s.t. } \boldsymbol{a}_v \geq \tau\}$. In the following, we set $\alpha = 0.5$ and $\pi = 75\%$ as suggested by (Meinshausen & Bühlmann, 2010), and use $l = 200$.

The randomized lasso does not fully address the problem of large groups of correlated variables. Indeed, the selection score will undoubtedly tend to zero as the size of group increase, and be confounded by selection scores of uninformative variables due to chance.

**Adding randomized clustering** To improve the stability of the estimation we propose to work with clusters of correlated variables. The motivation is that clustering variables and replacing groups of variables by their mean value will reduce the number of variables and improve the properties of the resulting design $\boldsymbol{X}_{\text{red}}$: less variables and less correlation between them. As the variables are voxels defined over a 3 dimensional grid and as correlation between voxels is strongly local, we can further constrain the clustering procedure to cluster only neighboring voxels. Following Michel et al. (2012), we use spatially-constrained Ward hierarchical clustering to cluster the voxels in $q$ spatially connected regions in which the signal is averaged. When merging two singleton clusters, the Ward criteria chooses the samples with largest cross-covariance. The benefit of this algorithm is thus that it is a bottom-up approach that captures well local correlations into spatial clusters. In addition, it gives a multi-scale representation: a tree of nested regions.

We run the clustering algorithm in the randomization loop of the randomized lasso: after sub-sampling the observations and rescaling the variables. As the hierarchical tree is estimated each time on a random fraction of the data, the tree is different for every randomization. Note that a similar randomization on trees by resampling is performed in the Random Forests algorithm (Breiman, 2001). It is motivated

**Algorithm 1** Randomized ward lasso

**Require: Input:** datasets $\boldsymbol{X} \in \mathbb{R}^{n \times p}$, target variable $\boldsymbol{y} \in \mathbb{R}^n$, number of clusters $q$, penalization $\lambda$, number of resampling $l$, scaling $\alpha \in [0, 1]$, sub-sampling fraction $\pi \in [0, 1]$
1: **for** $i = 1$ **to** $l$ **do**
2:  Sub-sample: $\tilde{\boldsymbol{X}} \leftarrow \boldsymbol{X}_J$, $\tilde{\boldsymbol{y}} \leftarrow \boldsymbol{y}_J$ where $J \subset \{1 \ldots n\}$, card$(J) = \pi n$, $\tilde{\boldsymbol{X}} \in \mathbb{R}^{\pi n \times p}$, $\tilde{\boldsymbol{y}} \in \mathbb{R}^{\pi n}$
3:  Randomize feature scaling: $\tilde{\boldsymbol{X}} \leftarrow \tilde{\boldsymbol{X}} \cdot \boldsymbol{w}$, where $\boldsymbol{w} \in \mathbb{R}^p$, $\boldsymbol{w}_i = 1 - \alpha \mu_i$, $\mu_i \sim \text{Bern}(0.5)$
4:  Cluster features, and use the clusters mean: $\tilde{\boldsymbol{X}}_{\text{red}} \leftarrow \text{ward}(\tilde{\boldsymbol{X}})$, $\tilde{\boldsymbol{X}}_{\text{red}} \in \mathbb{R}^{\pi n \times q}$
5:  Estimate $\boldsymbol{\beta}_{\text{red}}^i \in \mathbb{R}^q$ from $\tilde{\boldsymbol{X}}_{\text{red}}$ and $\tilde{\boldsymbol{y}}$ with sparse linear regression
6:  Label initial features with estimated coefficient on corresponding cluster: $\boldsymbol{\beta}^i \leftarrow \text{ward}^{-1}(\boldsymbol{\beta}_{\text{red}}^i)$, $\boldsymbol{\beta}^i \in \mathbb{R}^p$
7: **end for**
8: **return** stability scores $\boldsymbol{a}_v = \#\{k \text{ s.t. } v \in \text{supp}(\boldsymbol{\beta}^i)\}/l$, $\boldsymbol{a}_v \in [0, 1]$ by the high-variance of estimators based on greedy tree construction. Similarly, in *consensus clustering* (Monti et al., 2003) resampling is applied to the construction of clusters to account for the sensibility of the algorithms, that are most often based on greedy approaches to non-convex problems. After variable clustering, the sparse estimator can be fitted on a $q$-dimensional dataset. A variable is marked as selected in repetition $k$ if it belongs to a cluster with a non-zero weight. As the estimated supp$(\boldsymbol{\beta}^i)$ is in $\mathbb{R}^q$, we will write $v \in \text{supp}(\text{ward}^{-1}(\boldsymbol{\beta}^i))$, where ward$^{-1}$ amounts to assigning to the initial features the coefficient estimated for the center of the cluster that they belong to. The full estimation procedure, that we denote by *randomized ward lasso*, is summarized in Algorithm 1.

The main benefit of the additional clustering step is to reduce the correlations in the design matrix, therefore improving the behavior of sparse methods. Indeed, as correlated features are merged, the conditioning of sub-matrices of $\boldsymbol{X}_{\text{red}}$ improves, and thus the conditions for control of false detections with the randomized lasso are fulfilled for $q$ small enough. The key to the success of our method is this tight control: with high probability, clusters selected during each iteration all contain an original non zero feature. When back-projecting the selection frequency in the non-clustered space, because of the randomness of the cluster shapes in the different iterations, the selection frequency score can outline structures of a smaller size than the cluster size. There is an inherent trade-off in the choice of $q$: large clusters will most likely not match the geometry of the support of $\boldsymbol{\beta}$ but are more easily selected by sparse regression and are therefore more stable. Note that the separation between the signal and noise subspace may not be improved by the clustering, but it is not central to the success of the randomized lasso.

Importantly, thanks to the clustering, our method can select more variables than the number of observations, which would be impossible with a standard Lasso-type estimator and difficult with only randomization. Formally, the randomized ward lasso can be seen as applying a sparse estimator on random projections, learned by clustering. Importantly, those random projections are made of a small number of voxels, and thus are consistent with the assumed sparse representation of our problem. In addition, as they are not performed in rotationally-invariant directions, the resulting estimator is not rotationally invariant, and we suggest that it can achieve sub-linear sample complexity.

Finally, the randomized ward lasso is computationally tractable in the large $p$ settings. Indeed, agglomerative clustering can be implemented in linear time with regards to the number of possible agglomerations tested (Müllner, 2011). As we constrain clustering on the spatial grid, this number of agglomeration grows only linearly with $p$. The cost of a Lasso estimate on an $n \times p$ problem is $\mathcal{O}(n\,p\,k)$ if it selects $k$ non-zero coefficients. Thus, for $l$ resampling iterations $q = \kappa p$ clusters, $\kappa \in [0, 1]$, the computational complexity of the randomized ward lasso scales as $\mathcal{O}(l\,\kappa^2\,n\,p\,k)$. Importantly for our applications, the scaling in $p$ is linear. This is to be contrasted with sparse methods imposing a spatial regularization, such as group Lasso with overlapping spatial groups.

## 4. Experimental results

### 4.1. Synthetic data

**Data generation** We compare different sparse estimators on simulated data for which the ground truth is known. We simulate data according to a simple model reflecting the key elements of brain imaging data: local correlations of the design matrix $\boldsymbol{X}$, *i.e.* the observed brain images, and spatially clustered non-zero weights. Specifically, we work on a $(32 \times 64)$ grid of 2048 features. We generate an *i.i.d* Gaussian design $\boldsymbol{X}$ that we spatially smooth with a 2D Gaussian filter of standard deviation $\sigma$. The weight vectors $\boldsymbol{\beta}$ to recover have $k = 64$ non-zero coefficients. We split them into spatial clusters of size $c$, evenly-separated on the spa-tial grid. Non-zero coefficients are uniformly sampled between $\beta_{\min} = .2$ and $1 + \beta_{\min}$. Finally, the target variable $\boldsymbol{y}$ is generated according to the linear model with additive Gaussian $i.i.d$ noise. The amplitude of the noise is set to ensure that $\boldsymbol{X}$ explains 80% of the variance of $\boldsymbol{y}$ under the true model.

The two parameters of our procedure, the smoothing and the cluster size, control to which point the synthetic data violate the conditions for recovery. Sufficient smoothing can render arbitrarily ill conditioned groups of regressors located in a spatial neighborhood, thus violating sparse eigenvalue properties or the RIP. In particular, if the clusters are large, the design matrix restricted to the signal subspace $\boldsymbol{X}_S$ is ill conditioned, whereas if they are small, the non-zero coefficients are well separated and $\boldsymbol{X}_S$ is well conditioned. Finally, as the smoothing increases, so does the coupling between the signal and noise subspaces, although this coupling is less important for large clusters.

We investigate sample sizes $n = 128$ and $256$, which are standard numbers for fMRI datasets. Note that with no smoothing, elements of $\boldsymbol{X}$ are Gaussian $i.i.d$ distributed, with identity covariance, and $n_{\min} \approx 1000$.

**Success metrics** We compare the ability to recover the true support of different estimators. Each estimator yields a score for each variable. The higher the score, the more likely is the variable to be in the support. We use precision-recall curves to quantify the ability of each estimator to recover the non-zero coefficients as the threshold on the scores is varied. We summarize the precision-recall by the area under the curve (AUC). An AUC of .5 is chance, while 1 is perfect recovery: there exists a threshold such that active and non-active features are perfectly separated. In practice, we consider that an AUC above 0.95 is a near-perfect recovery, and an AUC above .75 as usable. To match the context of real data where ground truth is unknown, note that we report performance for automatically selected parameters, and not best possible performance, unlike *e.g.* Grave et al. (2011).

**Results** We investigated two non-convex approaches: a greedy algorithm, orthogonal matching pursuit (Mallat & Zhang, 1993), and an iterative approach, the adaptive Lasso (Zou, 2006). We do not report the results as they did not achieve useful recovery. This failure can be understood because of the low number of samples, below $n_{\min}$ in non smoothed design, and the fragility of non-convex estimators when working with correlated designs. In addition, we studied screening based on F-tests, the elastic net, the randomized lasso, and our contribution, the random-ized ward lasso. Each time, we set the regularization parameters using 6-fold cross-validation to maximize the explained variance of left out data. For elastic net, we also set by cross-validation the amount of $\ell_2$ regularization, and for randomized ward lasso, the number of clusters. Setting penalties by minimizing test error gives no guaranties on estimation error in terms of recovery, but alternative methods like information-based criterion (IC) such as BIC or AIC do not hold in theory for small-sample settings. In our simulations cross-validation was superior over IC.

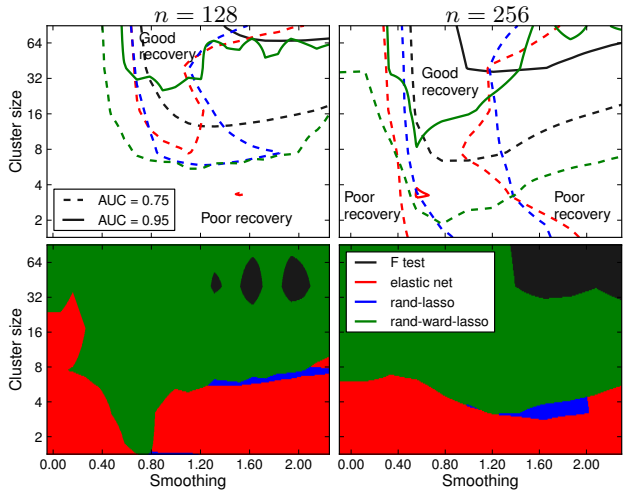

*Figure 1.* Recovery power on simulated data for $p = 2048$, $n = 128$ (left) or $n = 256$ (right) as a function of the cluster size and the smoothing. The recovery power is measured by the area under the precision-recall curve (AUC). **Top row**: contours at 0.95 and 0.75 of the AUC for different methods. **Bottom row**: best performing method.

*Figure 2.* Recovery on simulated data for $n = 128$ as a function of the clustering approach: learning clusters on the original $\boldsymbol{X}$, a random Gaussian $i.i.d$ signal, or our randomization approach.

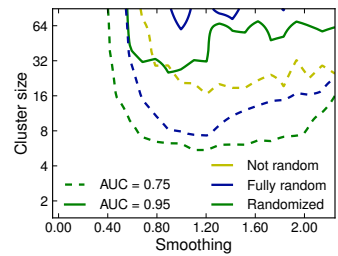

Results can be seen on figure 1. With no smoothing, all methods fail to give satisfactory recovery, although very large clusters can be partially recovered by randomized ward lasso with $n = 256$. For large cluster sizes, adding a small amount of correlation to the design improves the recovery. Indeed, the added correlation is consistent with the sparsity pattern to recover. Elastic net and randomized lasso can then recover even small clusters. Due to the small sample size, no method performs well in case of small clusters without any smoothing, nor in the case of small clusters and large smoothing, which corresponds to an ill-posed deconvolution problem. In the large cluster and large smoothing case, the univariate approach outperforms all others. In this situation, sub-matrices of the design matrix are too ill-conditioned for sparse methods to succeed, but the effective number of degrees of freedom to be estimated is strongly reduced by the inter-feature correlations: the univariate approach does not perform many independent hypothesis tests and thus the number of false positive is reduced. In a large region of the parameter space explored, for clusters composed of 8 or more features, and smoothing standard deviation of the order of a pixel, our proposed method, the randomized ward lasso, outperforms other methods. We note that in the small sample case ($n = 128$) parameter selection is particularly difficult for this method. Interestingly, we find that, in the settings that we explore, the randomized lasso is most often outperformed by the elastic net, suggesting that the elastic net's relaxation of the conditions on sparse eigenvalues is better suited to the problem than the randomized lasso's relaxation of the irrepresentable condition. Finally, we confirmed empirically that, in the randomized ward lasso, randomizing the clusters at each iteration was important: fully-random clusters outperform using a non-random clusters, and our randomized approach performs best (Fig. 2).

### 4.2. FMRI data

To assess the performance of our approach on real fMRI data, we investigated two datasets. The first one from Tom et al. (2007) is a gambling task where the subject is asked to accept or reject gambles that offered a 50/50 chance of gaining or losing money. Each gamble has an amount that can be used as target in a regression setting. The second dataset from Haxby et al. (2001) is a visual object recognition task. Each object category can then be used as target in a classification setting. In this setting, we use a sparse logistic regression model instead of a Lasso. We refer to Tom et al. (2007) and Haxby et al. (2001) for a detailed description of the experimental protocol. Data are publicly available on `http://openfmri.org`.

**Regression** After standard preprocessing (slice timing, motion correction, first level analysis with a general linear model, inter-subject spatial normalization), the regression dataset consists of 17 subjects with 48 fMRI observations per subject. Only the gain condition was used (see (Jimura & Poldrack, 2012)). The 48 observations contain 6 repetitions of the experiment with 8 levels of gain. FMRI volumes were downsampled to 4×4×4 mm voxels. When learning from the full set of 17 subjects, the training data therefore consist of 816 samples with approximately 26 000 voxels. The prediction here is inter-subject: the estimator learns on some subjects and predicts on left out subjects.

**Classification** In the object recognition experiment, 5 subjects were asked to recognize 8 different types of objects (*face*, *house*, *cat*, *shoe*, *scissors* etc.). We focus here on the binary classification task that consists in predicting whether the subject was viewing one object or the other (e.g. a *house* or a *face*). The data consist of 12 sessions, each with 9 volumes per category. When considering $N$ sessions for training, data consist of $18N$ volumes each containing about 30 000 voxels. The problem is here intra-subject: some sessions are used for learning and prediction is performed on left out sessions of the same subject.

As for such real data the ground truth is not known, we test these qualitative observations by training a predictive model on the voxels with the highest scores as in (Haury et al., 2011). The rationale is that if voxels with the highest F-scores contain more false positives than voxels with the highest stability scores then a predictive model trained with the first set should give significantly worse performance than a model trained with the second one. To validate this hypothesis, we train a $\ell_2$ logistic regression model on the best voxels as scored by the methods, for different object contrasts and numbers of training sessions. The number of selected voxels and the regularization parameter is optimized by cross-validation. For $T$ training sessions ($4 \leq T \leq 10$) we obtain a prediction score on $12-T$ left out sessions, unseen by the feature-selection method. For completeness, we also benchmark a linear SVM, as well as $\ell_1$ and $\ell_2$ logistic regressions trained on the whole brain. Experiments were performed with the scikit-learn (Pedregosa et al., 2011), using LIBLINEAR (Fan et al., 2008) for logistic regression and LIBSVM (Chang & Lin, 2011) for SVM.

**Results** Figure 3 shows maps of F values, the scores obtained with standard stability selection, and the scores obtained using our approach. The scores are not thresholded for visualization purposes, meaning that the zeros obtained are actually zeros. On both datasets, one can observe that despite being fairly noisy F-tests highlight well localized regions of the brain. Very similar regions are clearly outlined by our approach, whereas the standard stability selection procedure yields more spatially scattered selection scores. We also ran elastic-net in the regression settings (not displayed). The cross-validation lead setting a compromise between $\ell_2$ and $\ell_1$ penalty in favor of $\ell_1$: $\lambda(0.1\ell_2 + 0.9\ell_1)$. As a result, the maps were overly sparse, displaying single scattered voxels. Finally, in the regression and classification settings, we find $\approx$ 1000 and 2700 non-zero scores, which corresponds to a ratio $n/n_{\min}$ respectively $\approx$ 25 and 780 for $n_{\min} = 2p \log(p - k)$. While visual inspection of the figures suggests that thresholding the scores at 0 probably leads to false detections and thus over estimating $k$, the estimation problems tackled here can be safely considered as well beyond the reach of conventional sparse estimators in terms of number of samples.

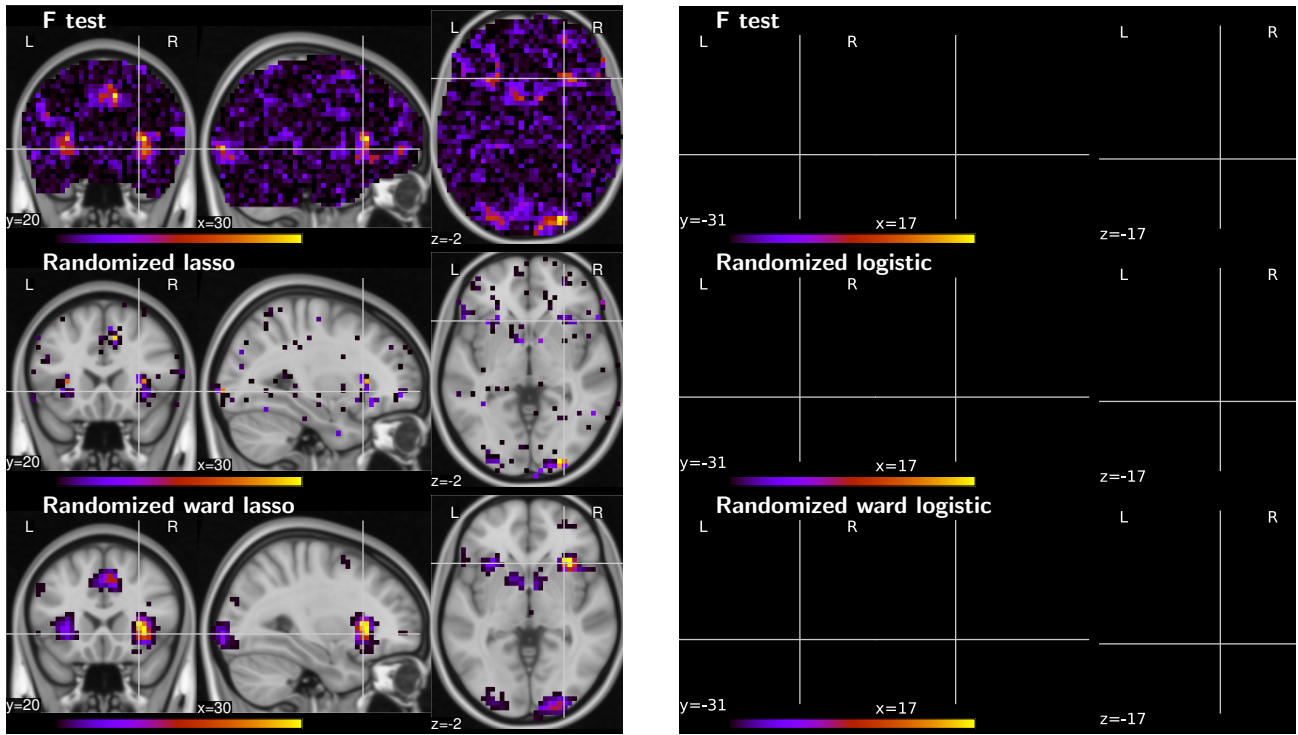

*Figure 3.* Score maps (unthresholded) as estimated by different methods on fMRI datasets. **Left column**: regression settings. **Right column**: classification settings (contrast *face* vs *house*) using 4 sessions of training data.

These prediction results are reported in Fig. 4. First we observe that a linear SVM and an $\ell_2$-logistic regression yield very similar results which is expected due to the similarity between the logistic loss and the hinge loss. Second, as this learning problem displays many non-informative features, using an $\ell_1$ penalization actually helps the prediction. Finally, we observe that using the best voxels as scored by the randomized ward logistic leads to significantly better performance than when using the voxels with the highest F-score. This suggests that it actually achieves a better identification of the voxels involved in the cognitive task.

## 5. Conclusion

Motivated by the problem of brain mapping using functional MRI data we address the problem of support identification when working with strongly spatially correlated designs as well as low number of samples. We propose to use a clustering of variables as well as randomization techniques, in order to position the problem in settings where a good recovery is achievable, *i.e.*, low correlation between variables in the support and limited number of variables compared to the number of samples. Further informed by the strong spatial correlations between voxels in fMRI data we constrain spatially the clusters. This constraint injects additional prior in the estimation to facilitate the recovery of spatially contiguous supports. Our formulation enables the use of computationally-efficient sparse linear regression models and yields an overall algorithmic complexity linear in the number of features.

Results on simulations highlight the settings in which our approach outperforms techniques like elastic net or univariate feature screening. Results on two publicly available fMRI datasets demonstrate how our approach is able to zero out some non-informative regions while outlining relevant clusters of voxels. Although the true support is unknown for real data, prediction scores obtained on the best voxels outperforms alternative supervised methods suggesting that our approach is not only better in terms of support identification but can also significantly improve prediction on such data.

The success of our approach with very small sample sizes lies in exploiting the characteristics of the prob-lem tackled: recovery of contiguous clusters from image measurements with local correlations. Indeed by clustering variables, we exploit the fact that their observed correlation shares common structure with the sparsity pattern to recover. Further work include investigating other clustering algorithms and comparing to more computationally costly but convex estimators such as using overlapping group penalties.

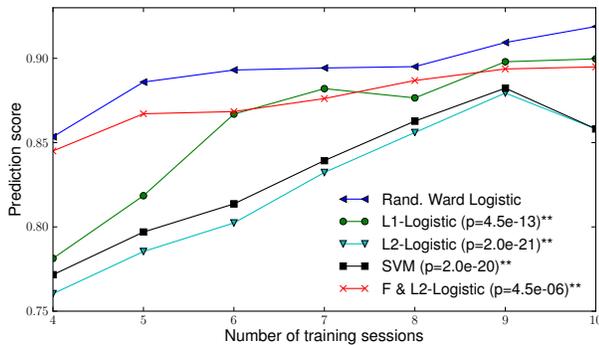

*Figure 4.* Average classification scores obtained using full brain linear SVM, $\ell_1$ and $\ell_2$-logistic regressions and $\ell_2$-logistic regression defined over the most significant voxels given by F-test (F & L2-Logistic) or randomized ward logistic. The average is computed over 6 binary classification tasks (face–house, face–cat, bottle–scissors, cat–house, shoe–house, shoe–face). P-values, computed with a Wilcoxon 1-sided paired, test if the randomized ward logistic prediction scores are larger than the ones obtained with the other methods. Randomized ward logistic significantly outperforms all the other methods.